\documentclass{article}
\usepackage{ijcai22}

\usepackage{times}
\usepackage{soul}
\usepackage{url}
\usepackage[hidelinks]{hyperref}
\usepackage[utf8]{inputenc}
\usepackage[small]{caption}
\usepackage{graphicx}
\usepackage{amsmath}
\usepackage{amsthm}
\usepackage{booktabs}
\usepackage{algorithm}
\usepackage{amssymb}

\usepackage{mathtools}
\usepackage{epsfig}
\usepackage{algpseudocode}
\usepackage{subfigure}

\title{Graph Partner Neural Networks for Semi-Supervised Learning on Graphs}

\author{
Langzhang Liang$^1$
\and
Cuiyun Gao$^1$\and
Shiyi Chen$^1$\and
Shishi Duan$^2$\and
Yu Pan$^1$\and
Junjin Zheng$^1$\and
Lei Wang$^2$\and
Zenglin Xu$^1$\footnote{The corresponding author.}
\affiliations
$^1$Harbin Institute of Technology, Shenzhen, China\\
$^2$NetEase\\
\emails
\{lazylzliang, chenshiyi.hit, iperryuu, zenglin\}@gmail.com,
gaocuiyun@hit.edu.cn,
\{duanshishi, hzwanglei2013\}@corp.netease.com,
zjj9342474@163.com
}

\begin{document}

\maketitle

\begin{abstract}
Graph Convolutional Networks (GCNs) are powerful for processing graph-structured data and have achieved state-of-the-art performance in several tasks such as node classification, link prediction, and graph classification. However, it is inevitable for deep GCNs to suffer from an over-smoothing issue that the representations of nodes will tend to be indistinguishable after repeated graph convolution operations. To address this problem, we propose the Graph Partner Neural Network (GPNN) which incorporates a de-parameterized GCN and a parameter-sharing MLP. We provide empirical and theoretical evidence to demonstrate the effectiveness of the proposed MLP partner on tackling over-smoothing while benefiting from appropriate smoothness. To further tackle over-smoothing and regulate the learning process, we introduce a well-designed consistency contrastive loss and Kullback–Leibler (KL) divergence loss. Besides, we present a graph enhancement technique to improve the overall quality of edges in graphs. While most GCNs can work with shallow architecture only, GPNN can obtain better results through increasing model depth. Experiments on various node classification tasks have demonstrated the state-of-the-art performance of GPNN. Meanwhile, extensive ablation studies are conducted to investigate the contributions of each component in tackling over-smoothing and improving performance.
\end{abstract}
\section{Introduction}
Recently, Graph Neural Networks (GNNs) have become a hot topic in deep learning for the potentials in modeling non-Euclidean data. Early works ~\cite{DBLP:conf/icml/ZhuGL03,DBLP:conf/nips/ZhouBLWS03,DBLP:journals/jmlr/BelkinNS06} adopt some form of explicit graph laplacian regularization, which restricts the flexibility and expressive power of GNNs. Later methods can be divided into two categories, namely the spectral-based methods and the spatial-based ones~\cite{DBLP:journals/tnn/WuPCLZY21}.
Graph Convolutional Network (GCN) ~\cite{DBLP:conf/iclr/KipfW17} bridges the above two categories via a first-order Chebyshev approximation of the spectral graph convolutions, which is similar to perform message passing operations on neighbor nodes.

\begin{figure}
    \centering
    \subfigure[Dirichlet energy of node embeddings, the squares of maximum and minimum singular values of weight matrix at the k-th layer of a 8-layer GCN on Cora dataset.] {\includegraphics[width=0.48\columnwidth,scale=1.2]{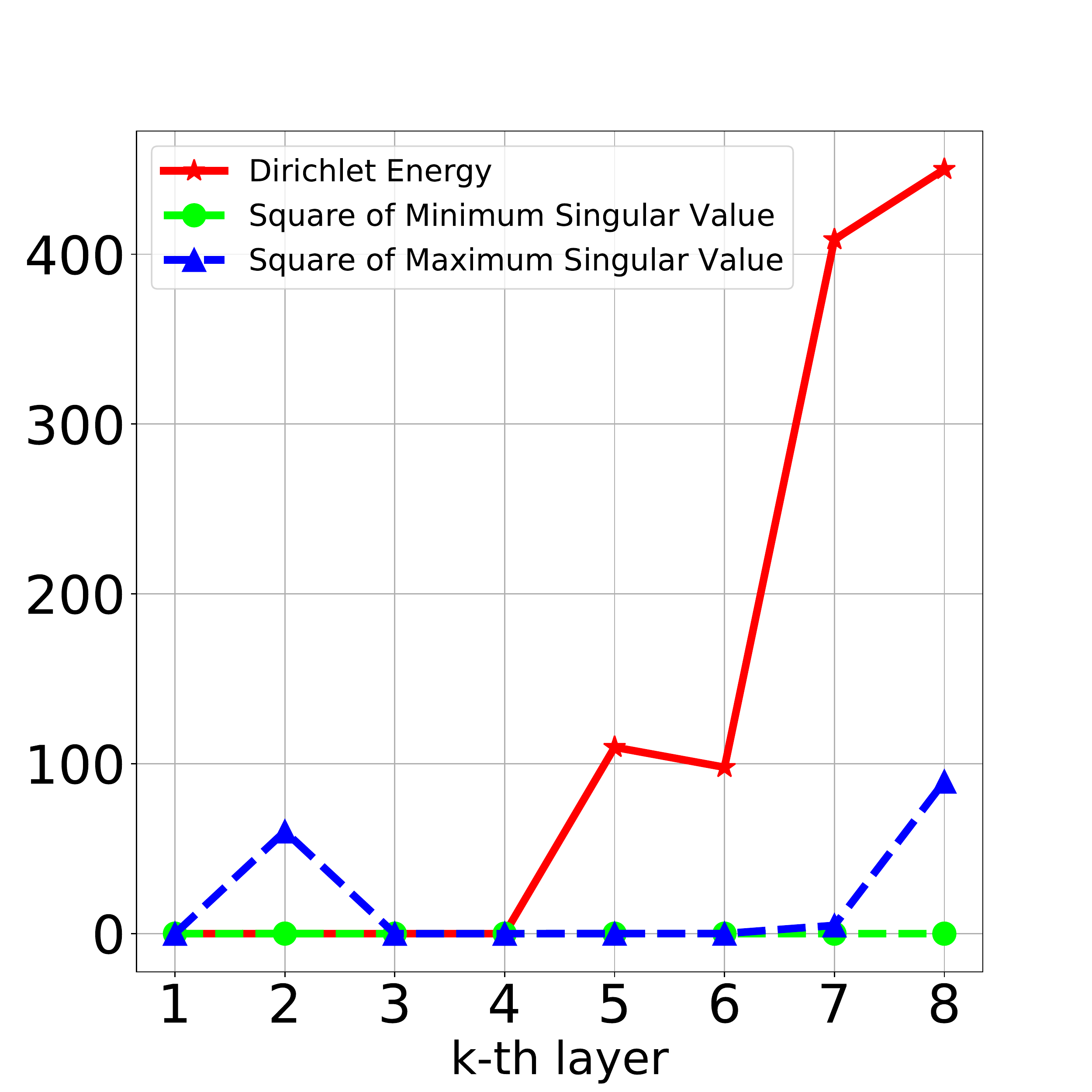}\label{}}
    \subfigure[visualization of node embeddings of the 4-th layer.]
    {\includegraphics[width=0.48\columnwidth,scale=1.2]{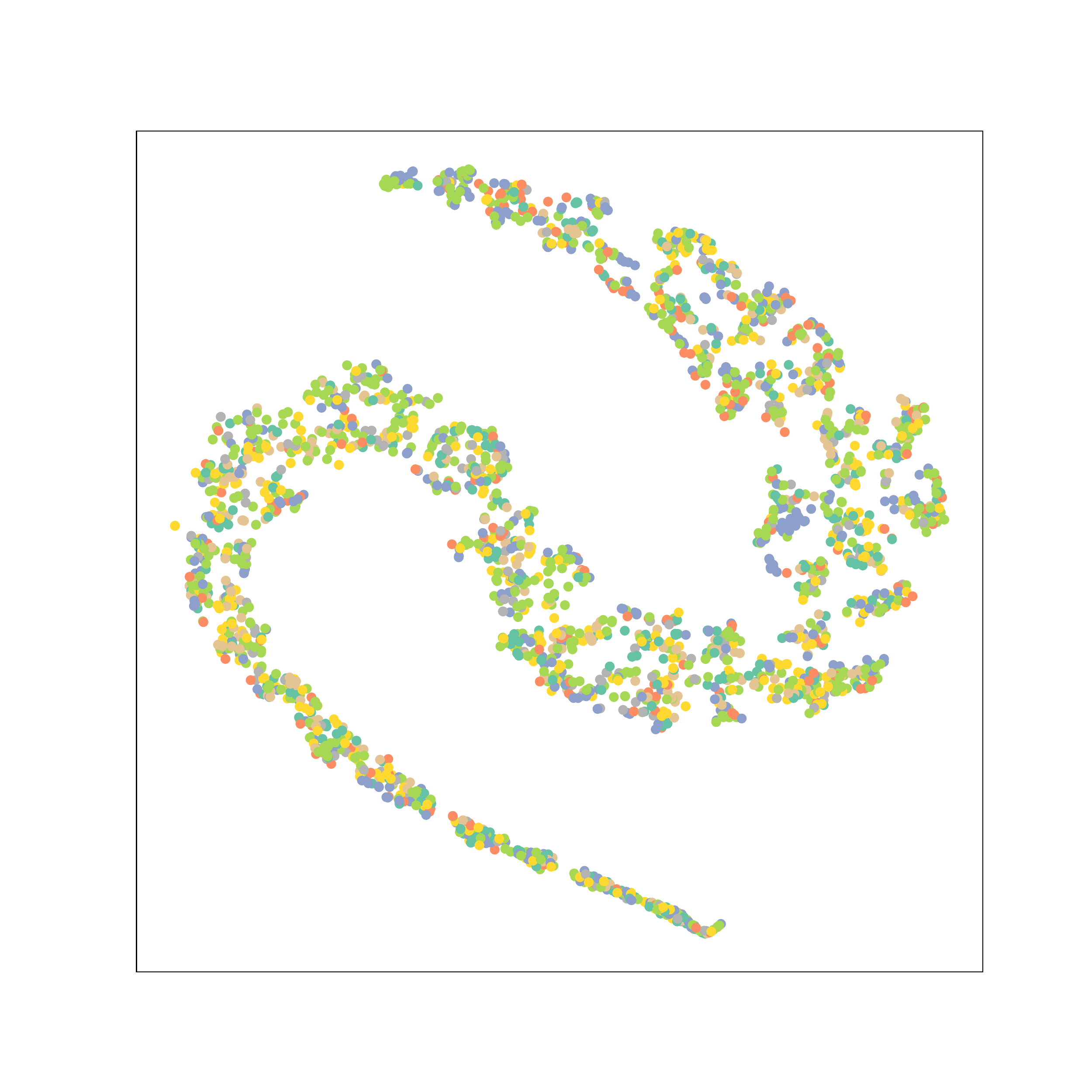}}
    \caption{Impact of feature transformation on over-smoothing. The Dirichlet energy at the first 4 layers is almost zero, which reveals that the pairwise distances are almost zero and the node embeddings are indistinguishable . We argue that the extremely small singular values of weight matrices give rise to this phenomenon.}
    \label{fg:intro}
\end{figure}

\begin{figure*}[t]
\centering
\includegraphics[width=1.9\columnwidth]{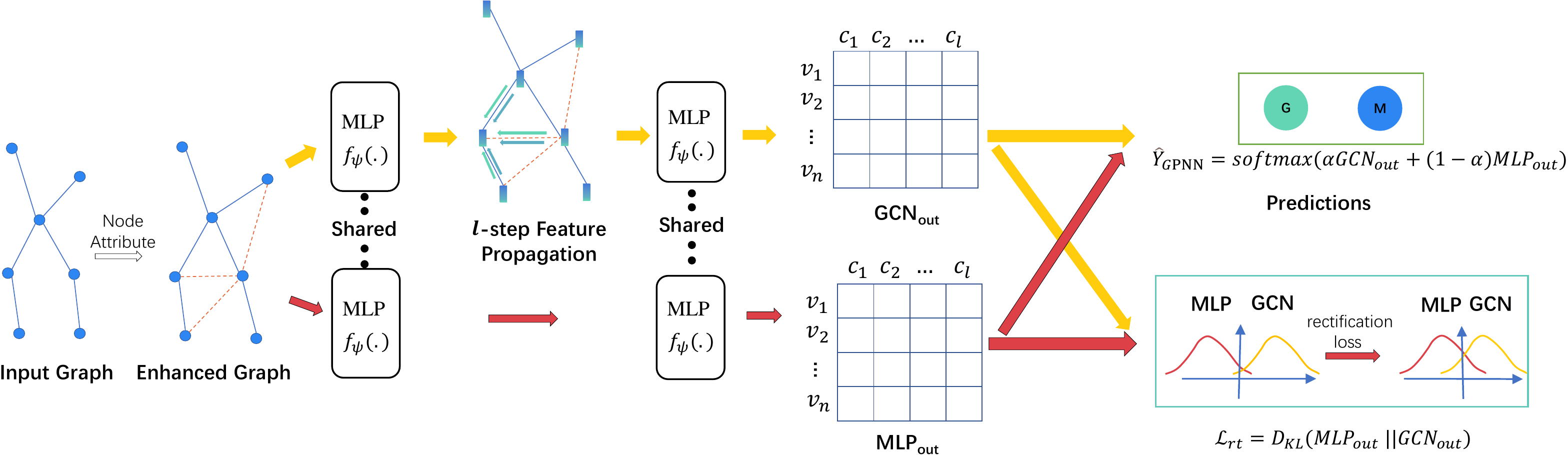} 
\caption{Overall architecture of the proposed model. We first perform graph enhancement to generate a new graph that incorporates the node attribute as supplemental information. The blue lines in the enhanced graph denote original edges in the graph, and the dotted lines represent newly added edges by graph enhancement. Then the input feature matrix, $\mathbf{X}$, is fed into the two branches of the model, respectively. The yellow lines represent the GCN branch which consists of a linear transformation along with an l-step feature propagation and another linear transformation. The Red lines represent the MLP branch including two 1-layer linear transformation. The two weight matrices of transformations are shared and trained jointly. Finally, the outputs of GCN and MLP are used to predict labels together and a rectification loss is calculated based on the KL divergence between them, which can be used to ensure consistency between the two branches. For clarify purposes, the contrastive loss is not depicted.}
\label{fig:model_framework}
\end{figure*}

Deep neural networks usually achieve better performance compared with shallow ones, but it is not the case in graph neural networks ~\cite{DBLP:conf/iclr/VelickovicFHLBH19}. Most GNNs obtain better results with a shallow structure, for example, GCN and GAT ~\cite{DBLP:conf/iclr/VelickovicCCRLB18} achieve their best performance at 2 layers. Repeated graph convolution operations can be approximated with a generalization of a Markov process, where the propagation matrix can be viewed as a generalized transition matrix. Notably, under certain conditions, a Markov process will exponentially converge to a steady state that is irrelevant to the initial state ~\cite{chung1997spectral}. In the context of our task, multiple graph convolution operations make the representations of nodes tend to be the same, \textit{i}.\textit{e}., indistinguishable and lose expressiveness, which is called over-smoothing. To tackle the over-smoothing issue in deep architectures, many approaches have been presented. Inspired by \cite{DBLP:conf/iclr/OonoS20}, DropEdge ~\cite{DBLP:conf/iclr/RongHXH20} suggests that over-smoothing can be relieved by randomly removing a fraction of edges from the input graph at each epoch. JKNet ~\cite{DBLP:conf/icml/XuLTSKJ18} finds it useful to reuse distinguishable features of the shallow layers, thus feeds previous layers' outputs into the last layer. Likewise, GCNII ~\cite{DBLP:conf/icml/ChenWHDL20} adopts the initial residual and identity mapping technique to retain initial information.

Although these methods have alleviated over-smoothing via randomizing the propagation or reusing information from previous layers, they ignore the impact of feature transformation, another key component of graph convolution layers, on over-smoothing. As shown a recent study, the convergence rate of over-smoothing is influenced by the singular values of weight matrices \cite{DBLP:conf/iclr/OonoS20}. We further observe that the stack of transformations causes the minimum singular values of weight matrices in deep GCNs to shrink to near zero, which leads to extremely low Dirichlet energy lower bounds of node representations, in other words, node representations become indistinguishable (Fg. \ref{fg:intro}). Removing the transformation is equivalent to adopt the fixed identity weight matrix, whose eigenvalues are all equal to 1. In this way, the convergence rate slows down. Unfortunately, decoupling feature transformation and propagation alone is not sufficient to extend GCN to a deep model. As the propagation step increases, node representations finally converge to a stationary point. 

Motivated by the above insights, we propose the Graph Partner Neural Network (GPNN) which builds GNNs with a de-parameterized GCN accompanied by an MLP partner. Instead of adding an MLP component directly, we employ a parameter-sharing MLP and further utilize a technique called consistency regularization to ensure the consistency between the outputs of MLP and GCN. The design principle of the proposed MLP partner is inspired by the Dirichlet energy restriction: with this partner, we establish controllable energy lower and upper bounds for GCN w.r.t the initial energy, which further enables GCN to learn discriminative yet smoothing representations. Besides, by decoupling the transformation and propagation, models are more flexible in learning from both attribute and graph structure, and the removal of transformation in GCN helps relieve over-smoothing.

Another obstacle to train deep GCNs is the inconsistency between the smoothing features and the initial features. Feature propagation can be viewed as a perturbation that brings the nodes to deviate from themselves via repeatedly aggregating neighbors. This perturbation may be so strong that the information of the node itself is totally covered by neighbors' information, then the over-smoothing occurs. Ensuring the consistency between the smoothing and initial feature enables the model  preserve more information on the node itself thus facilitating the training process of deep models. To this end, we design a consistency contrastive loss to maximize the agreement between the initial features and the smoothing features after the $l$-step propagation by encouraging the smoothing features to ``recall" the initial features. 

The existence of noisy connections, \textit{i}.\textit{e}., connected nodes belonging to different classes, is one key factor that aggravates the over-smoothing issue \cite{DBLP:conf/aaai/ChenLLLZS20}. Besides, the original graph may miss some high quality edges that are useful for diffusing training signals. Exploiting node attribute information is an effective way to enhance the quality of edges ~\cite{DBLP:conf/aaai/LiWZH18,DBLP:conf/iclr/DengZWZF20}.
We propose to enhance the connection quality by encoding both topology structure and node attribute information, which we call graph enhancement.

The main contributions are summarized as follows:
\begin{itemize}
\item We propose GPNN, a novel framework for training deep models, which is built with a de-parameterized GCN and a parameter-sharing MLP partner. Empirical and theoretical evidences show that the proposed MLP partner can enable GCN to benefit from smoothness and become robust to over-smoothing issue.
\item We design a consistency contrastive loss to facilitate the training of deep models. As we shall empirically demonstrate later via the Dirichlet energy, this loss can help learn distinguishable representations.
\item We propose a graph enhancement technique that can boost the graph with new highly-likely edges that have high quality compared with the original edges in terms of similarity and the same-label rate between connected nodes. 
\item Experimental results show that GPNN is powerful in solving the over-smoothing issue and successfully achieves the state-of-the-art performance on several node classification benchmarks.
\end{itemize}
\section{Problem Definition and Preliminaries}
\subsection{Problem Setup}
This paper focus on the semi-supervised node classification problem on graph-structured data. In detail, we define the graph-structured data as an undirected graph $\mathcal{G=(V,E)}$, where $\mathcal{V} = \{v_1, v_2, \dots, v_n\}$ is the node set of the graph $\mathcal{G}$, $\mathcal{E \subseteq V\times V}$ denotes the edge set where $e_{ij}=(v_i, v_j) \in \mathcal{E}$ represents an edge connecting the nodes $v_i$ and $v_j$. $\mathbf{X} \in \mathbb{R}^{n \times d}$ is the node feature matrix. The adjacency matrix is denoted as $\mathbf{A}\in \mathbb{R}^{n \times n}$, whose element $a_{ij} = 1$ represents that there is an edge between $v_i$ and $v_j$ and otherwise $a_{ij} = 0$. $c_i \in \{1,...,C\}$ is the label of $v_i$. $\mathcal{V}_l$ and $\mathcal{V}_u$ are the sets of labeled nodes and unlabeled nodes with $\left|\mathcal{V}_l\right|=n_l$ and $\left|\mathcal{V}_u\right|=n_u$, respectively. Given the label matrix $\mathbf{Y}_l \in \mathbb{R}^{n_l \times C}$ of labeled nodes, the task is to predict the label matrix of unlabeled nodes, $\mathbf{Y}_u \in \mathbb{R}^{n_u \times C}$.

\subsection{Graph Neural Networks} 
Graph Neural Networks generalize the deep learning techniques to process non-Euclidean data. Among them, Graph Convolutional Network \cite{DBLP:conf/iclr/KipfW17} is the most representative method, which reduces computation complexity via a first-order Chebyshev approximation of the spectral graph convolutions \cite{hammond2011wavelets,DBLP:conf/nips/DefferrardBV16} with the following layer-wise propagation rule
\begin{align}
    \mathbf H^{(l+1)} = \sigma(\tilde{\mathbf D}^{-\frac{1}{2}}\tilde{\mathbf A}\tilde{ \mathbf D}^{-\frac{1}{2}}\mathbf H^{(l)}\mathbf W^{(l)}).\label{eq:GNN_propagation}
\end{align}
Here, $\tilde{\mathbf D}^{-\frac{1}{2}}\tilde{\mathbf A}\tilde{\mathbf D}^{-\frac{1}{2}}$ is a propagation matrix with $\tilde{  \mathbf A}=\mathbf A+\mathbf I_n$, where $\mathbf I_n$ is an identity matrix and $\tilde{\mathbf D}_{ii} = \sum_j \tilde{\mathbf A}_{ij}$ is the degree matrix of $\tilde{\mathbf A}$. $\sigma(\cdot)$ denotes a non-linear function such as ReLU. 
GCN unifies the spectral domain and spatial domain, and
promotes subsequent works to focus on graph propagation, namely propagating information from each node to its neighbors with some deterministic propagation rules (\textit{i}.\textit{e}. message passing). 
From the perspective of message passing, GNNs can be generalized as 
\begin{equation}
   \mathbf H^{(l+1)} = \sigma(\text{AGGREGATE}(\mathbf H^{(l)}\mathbf W^{(l)},\mathbf A)).\label{eq:general_propagation}
\end{equation}
Here, given the representations of nodes $\mathbf H^{(l)}$ along with the adjacent relation matrix $\mathbf A$, $\text{AGGREGATE}(\cdot)$ represents an aggregation function that outputs representations $\mathbf H^{(l+1)}$ after neighbor aggregation. $\mathbf W^{(l)}$ is a learnable weight matrix at the $l^{th}$ layer, and $\mathbf H^{(0)} = \mathbf{X}$.

\subsection{Contrastive Learning}
The unsupervised contrastive learning is expected to provide extra supervised signals to guide the learning process of encoders. The main idea of contrastive learning is to learn representations such that positive sample pairs stay closer over negative pairs in the embedding space~\cite{khosla2020supervised}, which can be formulated as 
\begin{equation}
    s(f(\mathbf x_i),f(\mathbf x_i^+)) >> s(f(\mathbf x_i),f(\mathbf x_i^-)),\label{eq:cl_target}
\end{equation}
where $\mathbf x_i^+$ is a positive sample of $\mathbf x_i$, $\mathbf x_i^-$ is a negative sample of $\mathbf x_i$, $f(\cdot)$ is an encoder, and $s(\cdot)$ is a score function used to measure similarity of two samples. To achieve this, the loss function of contrastive learning is usually defined as 
\begin{equation}
    \mathcal{L}_{cs}=\text{max}(-\frac{1}{n} \sum_{i=1}^n( s(\mathbf x_i,  \mathbf x_i^+)-
    \frac{1}{\tau} \sum_{k \in \Omega} s(\mathbf x_i, \mathbf x_{i,k}^-)) + S,0),\label{eq:general_cl_loss}
\end{equation}
where $x_{i,k}^-$ denotes the $k^{th}$ negative sample of $x_i$, $\Omega$ denotes the negative sample set, and $S$ is a desired gap.
The main differences in the definition of contrastive loss lie in the way to sample positive and negative pairs, and the score function to measure their similarity.

\subsection{Dirichlet Energy}
Dirichlet energy is widely used in graph signal analysis as a metric for measuring the embeddings' smoothness. A smaller energy value is highly related to over-smoothing, while larger one reveals that the node embeddings are over-separating.~\cite{cai2020note}. Considering learned node embeddings $\mathbf{H}^{(l)} \in \mathbb R ^{n \times d}$ at the $l^{th}$ layer, the Dirichlet energy $E(\mathbf H^{(l)})$ is defined as
\begin{equation}
  \begin{split}
      E(\mathbf H ^{(l)}) &= tr(\mathbf H^{(l)^T} \widetilde{\mathbf L}\mathbf{H}^{(l)})\\
      &=\frac{1}{2} \sum a_{ij} \Vert \frac{h_i^l}{\sqrt{1+d_i}} - \frac{h_j^l}{\sqrt{1+d_j}} \Vert_2^2,
  \end{split}\label{eq:dirichlet_energy}
\end{equation}
where $tr(\cdot)$ is trace of a matrix, $\widetilde {\mathbf L} = \mathbf{I}_n-{\widetilde {\mathbf D}}^{-\frac{1}{2}}\widetilde {\mathbf A}{\widetilde {\mathbf D}}^{-\frac{1}{2}}$ is the augmented normalized Laplacian matrix, $a_{ij}$ is the $(i,j)^{th}$ element of adjacency matrix, and $d_i$ is the degree of node $i$. 
The lower bound of the Dirichlet energy at the $l^{th}$ layer is given by
\begin{equation}
    (1-\lambda_1)^2 s_\text{min}^{(l)} E(\mathbf H^{(l-1)}) \le E(\mathbf H^{(l)}) \le (1-\lambda_0)^2 s_\text{max}^{(l)} E(\mathbf H^{(l-1)}),
\label{eq:upper_bound}
\end{equation}
where $\lambda_0$ and $\lambda_1$ are the non-zero eigenvalues of $\widetilde{\mathbf{L}}$ that are most close to 0 and 1 respectively and locate within $[0,2)$. $s_\text{min}^{(l)}$ and $s_\text{max}^{(l)}$ are the squares of the minimum and maximum singular values of matrix $\mathbf W^{(l)}$, respectively (see \cite{zhou2021dirichlet} for proof).

\section{Related Work}
APPNP~\cite{DBLP:conf/iclr/KlicperaBG19} utilizes graph convolution and initial residual to approximate the personalized propagation while decoupling the transformation and propagation for fewer parameters. SGC~\cite{DBLP:conf/icml/WuSZFYW19} and LightGCN~\cite{DBLP:conf/sigir/0001DWLZ020} also decouple feature propagation and transformation for reducing computational complexity. Surprisingly, our experimental results clearly show that performing feature transformation in deep vanilla GCN gives rise to over-smoothing due to the extremely small singular values of the weight matrices. On the contrary, de-parameterization is equivalent to adopt the fixed identity weight matrix, whose singular values are all 1. To this end, we remove the transformation in graph convolution layers for controlling the singular values. The identity mapping technique used in GCNII~\cite{DBLP:conf/icml/ChenWHDL20} is similar to removing most of the parameters while preserving a small part of learning ability. 
\cite{DBLP:conf/kdd/0017ZB0SP20} propose to construct a k-nearest neighbor (kNN) graph by calculating the similarity between all potential node pairs. However, the k-nearest neighbor graph neglects the topological information and has $O(n^2)$ computational complexity. Our graph enhancement can exploit both node attribute and topology information and thus explore some new edges with high connection quality.

GraphMix~\cite{DBLP:conf/aaai/VermaQKLBKT21} utilizes a parameter-sharing MLP to facilitate the training process of GCN with interpolation-based data augmentation~\cite{DBLP:conf/iclr/ZhangCDL18}.
GCNII \cite{DBLP:conf/icml/ChenWHDL20} adopts initial residual to retain initial information. However, simply adding a fraction of initial features to each layer somewhat overwhelms the learned embeddings of later layers. GPNN only incorporates the initial features at the last layer, which is sufficient to prevent over-smoothing as we prove later.

Recent advances in multi-view visual representation learning~\cite{DBLP:conf/eccv/TianKI20,DBLP:conf/nips/BachmanHB19,DBLP:conf/icml/ChenK0H20} lead to a surge of interest in applying contrastive learning to graph data. Most of them utilize the data augmentation technique to generate multi views of a same graph then maximize the mutual information between different views~\cite{DBLP:conf/icml/HassaniA20,DBLP:conf/kdd/QiuCDZYDWT20,DBLP:conf/aaai/WanPY021}. In this work, contrastive learning is used to effectively alleviate over-smoothing by maximizing the agreement between initial features and aggregated embeddings.

\section{Methodology}

\subsection{Graph Enhancement}
Conventionally, message passing of GNN models takes place among direct (one-hop) neighbors. However, 
nodes connected with edges do not necessarily share the same label. Therefore, we derive an enhanced graph by encoding both the topology and node attribute information. Specifically, we first obtain an attribute graph by calculating the cosine similarity between two $k$-hop connected nodes based on the raw feature matrix $\mathbf{X}$. $\mathbf{A}^k\in \mathbb{R}^{n \times n}$ is the $k^{th}$ power of $\mathbf{A}$, and $a_{ij}^k \geq 1$ in $\mathbf A^{k}$ represents that $v_i$ can reach $v_j$ within $k$ steps. We define the edge weight as $w_{ij}=(x_{i} \cdot x_{j})/(\Vert  x_{i}\Vert \Vert x_{j}\Vert)$, where $ x_{i}$ and $x_{j}$ are the corresponding feature vectors of $v_i$ and $v_j$ in $\mathbf X$, respectively. Then, we denote the adjacency matrix of the attribute graph as $\mathbf{A}_{attr}$, whose elements are defined as
\begin{equation}
    a_{ij}^{attr}=
\begin{cases}
w_{ij},& a^k_{ij}\geq 1, w_{ij} \geq t\\
0,& \text{others}
\end{cases}, 1 \leq i,j \leq n,
\label{eq:enhenced_adjacency_element}
\end{equation}
where $t$ is a threshold value to control the connection density of the attribute graph. In practice, we fix $k=3$ across all experiments. 
Finally, we normalize the attribute adjacency matrix as $\tilde a_{ij}^{attr}= \frac{a_{ij}^{attr}}{\sum_j a_{ij}^{attr}}$, then construct an enhanced propagation matrix by combining the augmented normalized adjacency matrix and the normalized attribute adjacency matrix using weighted sum:
\begin{equation}
    \mathbf A_{en} = \beta \tilde{\mathbf D}^{-\frac{1}{2}}\tilde{\mathbf A}\tilde{ \mathbf  D}^{-\frac{1}{2}} + (1-\beta)\tilde{\mathbf A}_{attr}, \label{eq:enhenced_adjacency_matrix}
\end{equation}
where $0< \beta <1$ is a hyperparameter. We then remove the linear transformation and nonlinearity in the graph convolution layers. Therefore, the propagation rule used in GPNN is defined as
\begin{equation}
    \mathbf H^{(l+1)} = \mathbf A_{en}\mathbf H^{(l)}, \label{eq:gpnn_propagation}
\end{equation}
Notably, the enhanced graph enlarges the receptive field for nodes in each layer. For example, let $k$ be 3, with 16 propagation steps, a node can aggregate information from its neighbors up to $48$ hops.

\subsection{Graph Partner Neural Networks}
The proposed GPNN combines the MLP branch (with co-training and parameter-sharing) and the GCN branch to predict labels together, as  shown in Figure \ref{fig:model_framework}. This method emphasizes the importance of attribute information of the node itself than the vanilla GCN. Firstly, the output of the MLP branch will be directly optimized as it is a part of the label predictions; secondly, the MLP branch shares the feature transformation matrix with the GCN branch, thus MLP also affects the output of the GCN. In order  to obtain a lower-dimensional representation, we first apply a fully-connected layer and point-wise nonlinear activation function $\text{ReLU}= \text{max}(x,0)$ on the input feature matrix $\mathbf{X}$  as shown below,
\begin{equation}
    \mathbf H^{(1)}=\text{ReLU}(\mathbf X\mathbf W^{(1)}),
\label{eq:mlp_1}
\end{equation}
where $\mathbf{W}^{(1)}\in \mathbb{R}^{d \times d_{hid}}$ is a learnable weight matrix. Then, the representation $\mathbf H^{(1)}$ will be input into both the MLP branch and the GCN branch. The final output of the MLP branch is calculated as follows,
\begin{equation}
    \mathbf H^{\phi_1} = \mathbf H^{(1)}\mathbf W^{(2)},
\label{eq:mlp_2}
\end{equation}
where $\mathbf{W}^{(2)} \in \mathbb{R}^{d_{hid} \times C}$ is a learnable weight matrix. For the GCN branch, we apply the $l$-step feature propagation on $\mathbf H^{(1)}$ before mapping the dimension to the class number $C$. The output of the GCN branch can be formulated as
\begin{equation}
    \mathbf H^{(l+1)} = \mathbf A_{en}^l \mathbf H^{(1)}, \label{eq:gpnn_1}
\end{equation}
\begin{equation}
    \mathbf H^{\phi_2} = \mathbf H^{(l+1)}\mathbf W^{(2)}.
    \label{eq:gpnn_2}
\end{equation}
Note that the weight matrices $\mathbf W^{(1)}$ and $\mathbf W^{(2)}$ are shared by the MLP branch and the GCN branch. Finally, we use the weighted sum of the two outputs to obtain probability distribution over classes, \textit{i}.\textit{e}., 
\begin{equation}
    \mathbf O = \text{softmax}(\alpha \mathbf H^{\phi_1} + (1-\alpha)\mathbf H^{\phi_2}),\label{eq:output}
\end{equation}
where $0<\alpha<1$ is a weight assigned to the output of the MLP branch. Afterward, the cross-entropy loss can be adopted to penalize the differences between the network output and the labels of the originally labeled nodes as
\begin{equation}
    \mathcal{L}_{ce} = - \sum_{i=1}^{n_l} \sum_{j=1}^c \mathbf Y_{ij} \ln \mathbf O_{ij}.\label{eq:ce_loss}
\end{equation}

\subsection{Model Rectification}
To boost the two information processing branches (\textit{i}.\textit{e}., MLP and GCN) to learn from each other, we introduce a rectification loss to penalize the distance between the output probability distributions of the MLP branch and the GCN branch. We choose KL divergence to measure the distribution distance as done in~\cite{hinton2015distilling}. The proposed rectification loss is presented as
\begin{algorithm}[htb]
\caption{The proposed GPNN algorithm.}
\label{alg:Framwork} 
\textbf{Input: }
Adjacent matrix $\mathbf{A}$; 
feature matrix $\mathbf{X}$; 
label matrix $\mathbf{Y}$;
maximum number of iterations $\mathcal{T}$
\\
\textbf{Output: }
Predicted label of each unlabeled node in graph and trained model parameters.
\begin{algorithmic}[1]
\State Construct an enhanced propagation matrix $\mathbf A_{en}$ by encoding both graph topology and node attribute information based on Eqs.(\ref{eq:enhenced_adjacency_element}) and (\ref{eq:enhenced_adjacency_matrix});
\For{$i=1$ to $\mathcal{T}$}
\State Perform forward propagation to obtain MLP output $\mathbf H^{\phi_1}$ and GCN output $\mathbf H^{\phi_2}$;
\State Calculate contrastive loss $\mathcal{L}_{cs}$ based on Eqs.(\ref{eq:cs_loss}) and (\ref{eq:cs_loss2});
\State Calculate rectification loss $\mathcal{L}_{rt}$ based on Eq.(\ref{eq:rt_loss});
\State Use both $\mathbf H^{\phi_1}$ and $\mathbf H^{\phi_2}$ to predict labels jointly with Eq.(\ref{eq:output}), then calculate cross-entropy $\mathcal{L}_{ce}$ based on Eq.(\ref{eq:ce_loss});
\State Update network parameters according to the overall objective function $\mathcal{L}$ in Eq.(\ref{eq:overall_loss});
\EndFor
\State Conduct label predictions with trained model.
\end{algorithmic}
\end{algorithm}
\begin{equation}
  \begin{split}
      \mathcal{L}_{rt} &= D_{KL}(\sigma(\frac{\mathbf H^{\phi_1}}{T})\Vert \sigma(\frac{\mathbf H^{\phi_2}}{T}))\cdot T^2\\
      &= \frac{1}{n} \frac{1}{C} \sum_{i=1}^n \sum_{j=1}^C p_{ij}(\log p_{ij}- \log q_{ij}) \cdot T^2, \label{eq:rt_loss}
  \end{split}
\end{equation}
where $\sigma(\cdot)$ stands for the softmax function, $T$ is a temperature, $p_{ij} = \frac{e^{\mathbf H^{\phi_1}_{ij}/T}}{\sum_{k=1}^C e^{\mathbf H^{\phi_1}_{ik}/T}}$ and $q_{ij} = \frac{e^{\mathbf H^{\phi_2}_{ij}/T}}{\sum_{k=1}^C e^{\mathbf H^{\phi_2}_{ik}/T}}$ denote the soften probability that node $v_i$ belongs to class $c_j$ predicted by the MLP branch and the GCN branch, respectively. A larger value of $T$ produces a ``softer" probability distribution over classes while a lower  value of $T$ leads to less attention to logits that are smaller than average. In practice, we simply set $T=2$ across all experiments. Since the magnitude of the gradient of the rectification loss scales as $1/T^2$, we multiply the loss by $T^2$ so that the relative contribution of rectification loss keeps unchanged if the temperature $T$ is changed.

\subsection{Consistency Contrastive Loss}
To ensure the node features after repeated aggregations are still consistent with the original node features, we design the following contrastive loss
\begin{equation}
    \mathcal{L}_{cs}=\text{max}(S - Score,0),\label{eq:cs_loss}
\end{equation}
\begin{equation}
    Score = \frac{1}{n} \sum_{i=1}^n( s( h_i^{(1)}, h_i^{(l+1)})-
    \frac{1}{\tau} \sum_{k \in \Omega} s(h_i^{(1)},h_k^{(l+1)})),\label{eq:cs_loss2}
\end{equation}
where $h_i^{(1)}$ and $h_i^{(l+1)}$ are the embeddings of $v_i$ in $\mathbf{H}^{(1)}$ and $\mathbf{H}^{(l+1)}$, respectively. We treat them as a positive pair. $h_k^{(l+1)}$ denotes the embedding of $v_k$ in $\mathbf{H}^{(l+1)}$, which is regarded as a negative sample of $h_i^{(1)}$, $S$ is a hyperparameter and $\Omega$ is a set of negative nodes randomly sampled over all embeddings in $\mathbf{H}^{(l+1)}$ for $\tau$ times. We define the score function $s(\cdot,\cdot)$as inner product, \textit{i}.\textit{e}., $s( h_i^{(1)}, h_i^{(l+1)}) = {h_i^{(1)}}^T\cdot h_i^{(l+1)}$.

With this loss, the model can identify the initial node features even given a corresponding feature after multi-step feature propagation. 
We find that using the inner product followed by softmax function as score function harms the model's performance, which may be attributed to the hard restriction as the cross-entropy loss encourages the probability of positive sample to approach 1. Besides, using cosine similarity as score function and neighbors as positive samples has similar effects but it needs higher complexity. We solely apply this cosine-based loss on Citeseer dataset since it yields significant improvement.

\subsection{Model Training}
Combining $\mathcal{L}_{ce}$ with the contrastive loss $\mathcal{L}_{cs}$ and the rectification loss $\mathcal{L}_{rt}$, the overall objective function of GPNN is presented as
\begin{equation}
  \mathcal{L} = \mathcal{L}_{ce} + \lambda_{rt} \mathcal{L}_{rt} + \lambda_{cs} \mathcal{L}_{cs}, \label{eq:overall_loss}
\end{equation}
where $\lambda_{rt}>0$ and $\lambda_{cs}>0$ are hyperparameters to weight the importance of $\lambda_{rt}$ and $\lambda_{cs}$, respectively. The detailed description of GPNN is provided in Algorithm \ref{alg:Framwork}
\section{Energy Analysis}
In this section, we theoretically prove that the proposed MLP partner can boost the representation learning with appropriate node smoothness while preventing over-smoothing. For simplicity and easy generalization, we use the standard propagation matrix $\mathbf P=\tilde{\mathbf D}^{-\frac{1}{2}}\tilde{\mathbf A}\tilde{\mathbf D}^{-\frac{1}{2}}$ as most GCN-based methods done.
\newtheorem{prop}{Proposition}
\begin{prop}
The MLP partner raises the Dirichlet energy of the last layer's embeddings, \textit{i}.\textit{e}.,
\begin{equation}
    E(\alpha \mathbf{H}^{\phi_1} + (1-\alpha) \mathbf{H}^{\phi_2} ) \ge E(\mathbf{H}^{\phi_2}),
\end{equation} \label{pro1}
\end{prop}
\proof We start by simplifying the notations with $\mathbf{H}^{\phi_1}\coloneqq \mathbf X$ and $\mathbf{H}^{\phi_2}\coloneqq \mathbf Y$.
\begin{equation}
  \begin{split}
      &E(\alpha \mathbf X + (1-\alpha) \mathbf Y) \\
      &= tr((\alpha \mathbf X + (1-\alpha) \mathbf Y)^T \widetilde{\mathbf L} (\alpha \mathbf X + (1-\alpha) \mathbf Y))\\
      &= tr(\alpha^2 \mathbf X^T \widetilde{\mathbf L} \mathbf X) + tr((1-\alpha)^2\mathbf Y^T \widetilde{\mathbf L} \mathbf{Y}) + 2tr(\alpha(1-\alpha) \mathbf X^T \widetilde{\mathbf L} \mathbf Y)\\
      &=\alpha ^2 E(\mathbf X) + (1-\alpha)^2E(\mathbf Y) + 2 \alpha (1-\alpha)tr(\mathbf X^T \widetilde{\mathbf L} \mathbf Y),\label{eq:energy21}
  \end{split}
\end{equation}

Recall that $\mathbf Y = \mathbf P^l \mathbf X$ and $\widetilde{\mathbf L} = \mathbf I_n - \mathbf P$, we can easily obtain $E(\mathbf{X}) \ge E(\mathbf{Y})$:
\begin{equation}
    \begin{split}
        E(\mathbf Y) &= tr((\mathbf P^l \mathbf X)^T  \widetilde{\mathbf L} (\mathbf P^l \mathbf X))\\
        &=tr(\mathbf X^T \mathbf P^l \widetilde{\mathbf L} \mathbf P^l \mathbf X)\\
        &=tr(\mathbf X^T  (\mathbf I_n - \widetilde{\mathbf L})^l \widetilde{\mathbf L} (\mathbf I_n - \widetilde{\mathbf L})^l \mathbf X )\\
        &=tr(\mathbf X^T  \widetilde{\mathbf L} (\mathbf I_n - \widetilde{\mathbf L})^{2l} \mathbf X )\\
        &\le(1-\lambda_0)^{2l} tr(\mathbf X^T  \widetilde{\mathbf L}  \mathbf X)\\
        &\le tr(\mathbf X^T  \widetilde{\mathbf L}  \mathbf X) = E(\mathbf X),
    \end{split}
\end{equation}

Then we have
\begin{equation}
    \begin{split}
        &E(\alpha \mathbf X + (1-\alpha) \mathbf Y) - E(\mathbf Y)\\
        & = \alpha ^2 E(\mathbf X) + 2 \alpha (1-\alpha)tr(\mathbf X^T \widetilde{\mathbf L} \mathbf Y) - \alpha (2-\alpha) E(\mathbf Y)\\
        &\ge \alpha^2 E(\mathbf{Y}) + 2 \alpha (1-\alpha)tr(\mathbf X^T \widetilde{\mathbf L} \mathbf Y) - \alpha (2-\alpha) E(\mathbf Y)\\
        &= 2 \alpha (1-\alpha)tr(\mathbf X^T \widetilde{\mathbf L}\mathbf Y) - 2 \alpha (1-\alpha) E(\mathbf Y)
    \end{split}
\end{equation}

Similarly, we derive $E(\mathbf Y) \le tr(\mathbf X^T \widetilde{\mathbf L}  \mathbf Y)$ as below.
\begin{equation}
    \begin{split}
        E(\mathbf Y) &= tr((\mathbf P^l \mathbf X)^T  \widetilde{\mathbf L} (\mathbf P^l \mathbf X))\\
        &=tr(\mathbf X^T \mathbf P^l \widetilde{\mathbf L} \mathbf P^l \mathbf X)\\
        &=tr(\mathbf X^T  (\mathbf I_n - \widetilde{\mathbf L})^l \widetilde{\mathbf L} \mathbf P^l \mathbf X)\\
        &\le(1-\lambda_0)^l tr(\mathbf X^T  \widetilde{\mathbf L} \mathbf P^l \mathbf X)\\
        &\le tr(\mathbf X^T  \widetilde{\mathbf L} \mathbf P^l \mathbf X)=tr(\mathbf X^T \widetilde{\mathbf L}  \mathbf Y),
    \end{split}
\end{equation}

Finally, we have
\begin{equation}
    E(\alpha \mathbf X + (1-\alpha) \mathbf Y ) -  E(\mathbf Y) \ge 0
\end{equation}

\begin{prop}
GPNN has theoretical energy lower and upper bounds w.r.t the initial energy $E(\mathbf{H}^{\phi_1})$ as follows:
\begin{equation}
    \alpha ^2 E(\mathbf{H}^{\phi_1} )\le E(\alpha \mathbf{H}^{\phi_1} + (1-\alpha) \mathbf{H}^{\phi_2} )\le E(\mathbf{H}^{\phi_1})[\alpha^2 + (1-\lambda_0)^l]
\end{equation}
\end{prop}

\proof The lower bound can be immediately derived with Eq.\ref{eq:energy21}, thus we omit the corresponding proof. According to Eq.\ref{eq:energy21} and $E(\mathbf Y) \le tr(\mathbf X^T \widetilde{\mathbf L}  \mathbf Y)$, we have
\begin{equation}
    \begin{split}
        &E(\alpha \mathbf X + (1-\alpha) \mathbf Y) \\
        &=\alpha ^2 E(\mathbf X) + (1-\alpha)^2E(\mathbf Y) + 2 \alpha (1-\alpha)tr(\mathbf X^T \widetilde{\mathbf L} \mathbf Y)\\
        &\le \alpha ^2 E(\mathbf X) +(1-\alpha^2)tr(\mathbf X^T \widetilde{\mathbf L} \mathbf Y)\\
        &=\alpha ^2 E(\mathbf X) + (1-\alpha^2) tr(\mathbf X^T  \widetilde{\mathbf L} \mathbf P^l \mathbf X)\\
        &\le \alpha ^2 E(\mathbf X) + (1-\alpha^2)(1-\lambda_0)^l tr(\mathbf{X}^T \widetilde{\mathbf L} \mathbf{X})\\
        &=\alpha ^2 E(\mathbf X) + (1-\alpha^2)(1-\lambda_0)^l E(\mathbf X)\\
        &\le E(\mathbf X)[\alpha^2 + (1-\lambda_0)^l]
    \end{split}
\end{equation}

We note that nodes have not aggregated any information from neighbors in $\mathbf{H}^{\phi_1}$, thus its energy can be viewed as the initial energy and is always large. An reasonable energy at the last layer should be significantly smaller than the initial energy since it indicates enough smoothness between node pairs, which is the essential of GNNs. However, an energy value that is close to zero reveals that the model is seriously suffering from over-smoothing issue.
The energy lower and upper bounds at the last layer of GPNN clearly show that by appropriately choosing $\alpha$ and $l$, GPNN can benefit from smoothness while preventing over-smoothing.
\section{Experiments}

\begin{table*}[]
\centering
\caption{Summary of classification accuracy results on Cora, CiteSeer, PubMed, Coauthor CS, Amazon Computers (A-Computers), and Amazon Photo (A-Photo). The results we report are the average accuracy associated with the standard deviation after 100 and 10 independent runs for the three citation networks and the other three datasets, respectively.}
\label{tab:node_classification_result}
\begin{tabular}{c|cccccc}
\midrule
\midrule
 Method   &Cora &CiteSeer &PubMed  & Coauthor CS & A-Computers & A-Photo  \\
\midrule
GCN                 &81.5 &70.3  &79.0                                  & 91.1 ± 0.5  & 76.3 ± 0.5         & 87.3 ± 1.0        \\
GAT            &83.0 ± 0.7 &72.5 ± 0.7  &79.0 ± 0.3                                       & 90.8 ± 0.8    & 79.8 ± 1.2         & 86.5 ± 1.3       \\
DGI         &81.7 ± 0.6 &71.5 ± 0.7 &77.3 ± 0.6                                        & 90.0 ± 0.3    & 75.9 ± 0.6         & 83.1 ± 0.5       \\
MVGRL     &82.9 ± 0.7  &72.6 ± 0.7  &79.4 ± 0.3                                          & 91.3 ± 0.1    & 79.0 ± 0.6         & 87.3 ± 0.3       \\
GRACE     &80.0 ± 0.4 &71.7 ± 0.6        &79.5 ± 1.1                                    & 90.1 ± 0.8    & 71.8 ± 0.4         & 81.8 ± 1.0       \\
JKNet  &81.8 ± 0.7  &68.1 ± 1.1 &78.8 ± 0.5 &90.5 ± 0.3   &80.2 ± 0.6  &87.5 ± 0.6  \\
APPNP &83.6 ± 0.5 &71.6 ± 0.4 &79.7 ± 0.5  &91.6 ± 1.2  &81.9 ± 0.6 &90.8 ±  0.8\\
JKNet(Drop)  &83.4 ± 0.6 &70.9 ± 0.6 &79.0 ± 1.0     &91.2 ± 0.4  &80.4 ± 0.5  &90.5 ± 0.4 \\
GCNII    &85.5 ± 0.5 &73.4 ± 0.6 & 80.2 ± 0.4                                            & 92.5 ± 0.4  & 80.8 ± 1.4        & 90.6 ± 0.4      \\
GPNN      &\textbf{86.0 ± 0.4}  &\textbf{76.5 ± 0.5}    &\textbf{80.8 ± 0.4} &\textbf{93.1 ± 0.5}  &\textbf{84.8 ± 0.5 }  &\textbf{91.6 ± 0.6}   \\ 
\midrule
\end{tabular}
\end{table*} 

\begin{table}[t]
\centering
\caption{Dataset statistics.}
\label{tab:dataset}
\begin{tabular}{c|cccc}
\midrule
\midrule
Datasets   & Classes          & Features     & Nodes    & Edges        \\ \midrule
Cora    & 7          & 1,433         & 2,708          &5,429             \\
CiteSeer      & 6          & 3,703          & 3,327   &4,732                 \\
PubMed      & 3            & 500          & 19,717     &44,338                \\
Coauthor CS & 15         & 6,805         & 18,333       &81,894        \\
A-Computers    & 10          & 767         & 13,752 & 245,861  \\ 
A-Photo      &8        &745      &7,650      &119,081 \\
\midrule
\end{tabular}
\end{table}

\subsection{Experimental Setup}
In this section, we evaluate the performance of GPNN against the state-of-the-art graph neural network models on six graph benchmark datasets.

\paragraph{Datasets} 
We use three citation networks Cora, CiteSeer and PubMed~\cite{sen2008collective}, one co-author network Coauthor-CS~\cite{shchur2018pitfalls}, and two Amazon product co-purchase networks Amazon Computers (A-Computers) and Amazon Photo (A-Photo)~\cite{shchur2018pitfalls} for experiments. The statistics of the datasets are shown in Table \ref{tab:dataset}. For the three citation networks, we follow the standard  training/validation/testing set split~\cite{DBLP:conf/icml/YangCS16}, with fixed 20 nodes per class for training, 500 nodes for validation and 1,000 nodes for testing. 
Since no public split is available for the other three datasets, we randomly choose 20 nodes per class for training, 30 nodes per class for validation, and the rest for testing.

\paragraph{Baselines}

We compare GPNN with a variety of methods, including two state-of-the-art shallow models: GCN~\cite{DBLP:conf/iclr/KipfW17}, GAT~\cite{DBLP:conf/iclr/VelickovicCCRLB18}, four recent deep models: JKNet~\cite{DBLP:conf/icml/XuLTSKJ18},  GCNII~\cite{DBLP:conf/icml/ChenWHDL20}, JKNet equipped with DropEdge~\cite{DBLP:conf/iclr/RongHXH20}, and APPNP~\cite{klicpera2018predict}. We also include three state-of-the-art GCN-based contrastive models: DGI~\cite{DBLP:conf/iclr/VelickovicFHLBH19}, GRACE~\cite{zhu2020deep} and MVGRL~\cite{DBLP:conf/icml/HassaniA20}.

\subsection{Node Classification Results}

For GCN, GAT, DGI, GRACE, and MVGRL, the results are directly taken from \cite{DBLP:conf/aaai/WanPY021}. For GCNII, we use the best results reported in their paper \cite{DBLP:conf/icml/ChenWHDL20} for the three citation networks (Cora/Citeseer/Pubmed). For GPNN and other baselines, we use grid search to choose hyperparameters based on the average accuracy on the validation set. 
Table~\ref{tab:node_classification_result} reports the node classification results on the benchmarks. The results demonstrate that GPNN successfully outperforms all the baselines across all six datasets. Especially, GPNN achieves new state-of-the-art performance on Citeseer dataset with 48 layers and receptive fields up to 144-hop neighbors. The experimental results show that by using the proposed MLP Partner and the consistency contrastive loss, one can significantly improve GCN's performance and effectively relieve over-smoothing. An interesting point we find is that the performance improvements of GPNN over GCNII on Citeseer, Amazon-Computers, and Amazon-Photo are relatively higher than that on other datasets, and only these three datasets contain isolated nodes. This result again verifies the claim that GPNN can flexibly learn from attribute and structure according to specific datasets.

\subsection{Over-Smoothing Analysis}

Due to over-smoothing, GCN~\cite{DBLP:conf/iclr/KipfW17} cannot benefit from a deep architecture as traditional neural networks do. GCN achieves its best performance at 2-layer. However, its performance will degrade when depth grows since the node representations will become indistinguishable gradually. Here, we study the robustness of GPNN to this issue via MADGap  \cite{DBLP:conf/aaai/ChenLLLZS20}, an overall quantitative measure of the smoothness of learned representations. A smaller value of MADGap denotes a shorter average cosine distance among all node pairs, which indicates a more severe over-smoothing problem. In this experiment, we evaluate GPNN, GCN, and GAT in terms of the MADGap values and classification performance w.r.t. different propagation steps(ranging from 2 to 15).
\begin{figure}
    \centering
    \subfigure[MADGap]{\includegraphics[width=0.48\columnwidth]{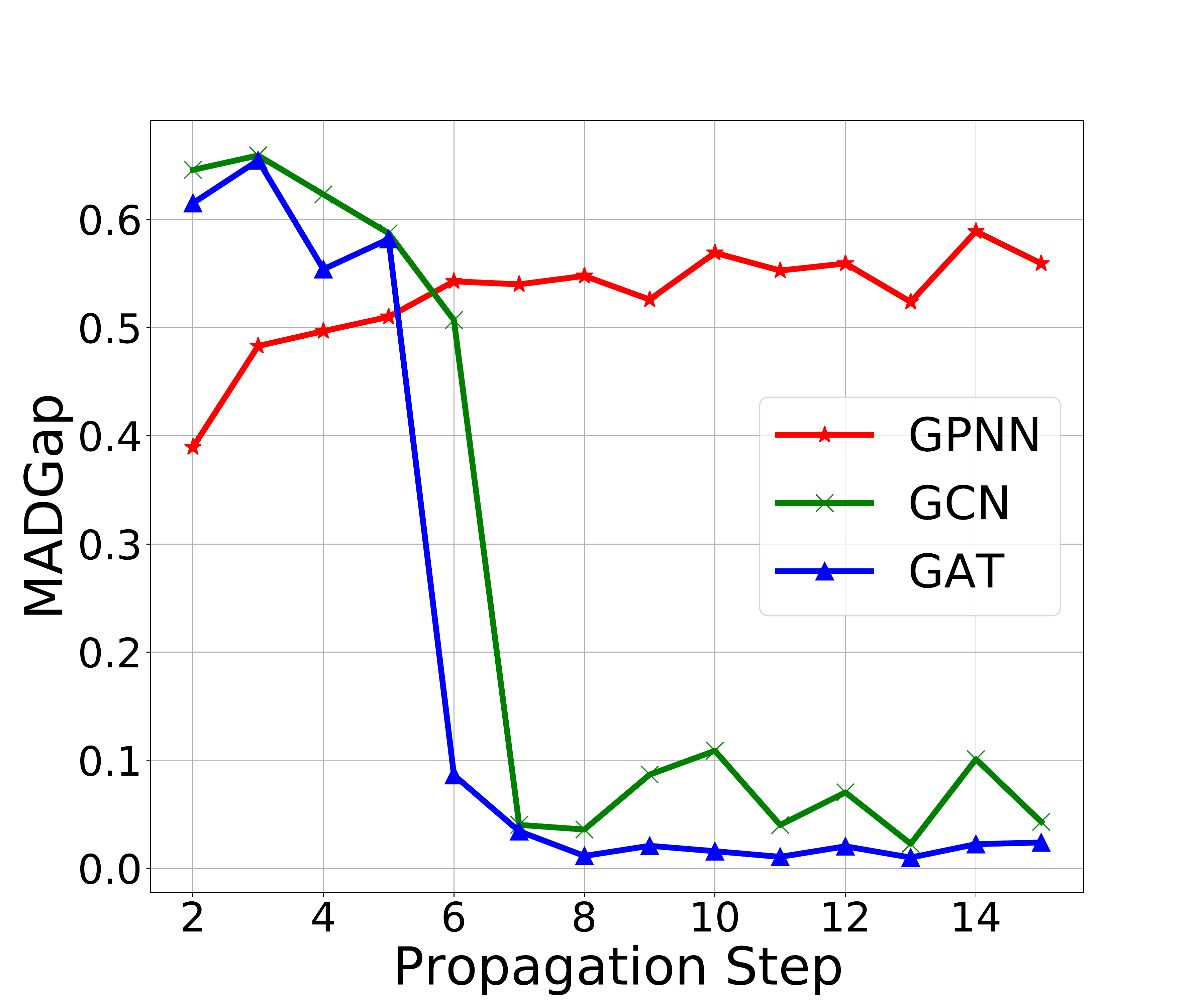}\label{subfig:oversmothing_madgap}}
    \subfigure[Accuracy]{\includegraphics[width=0.48\columnwidth]{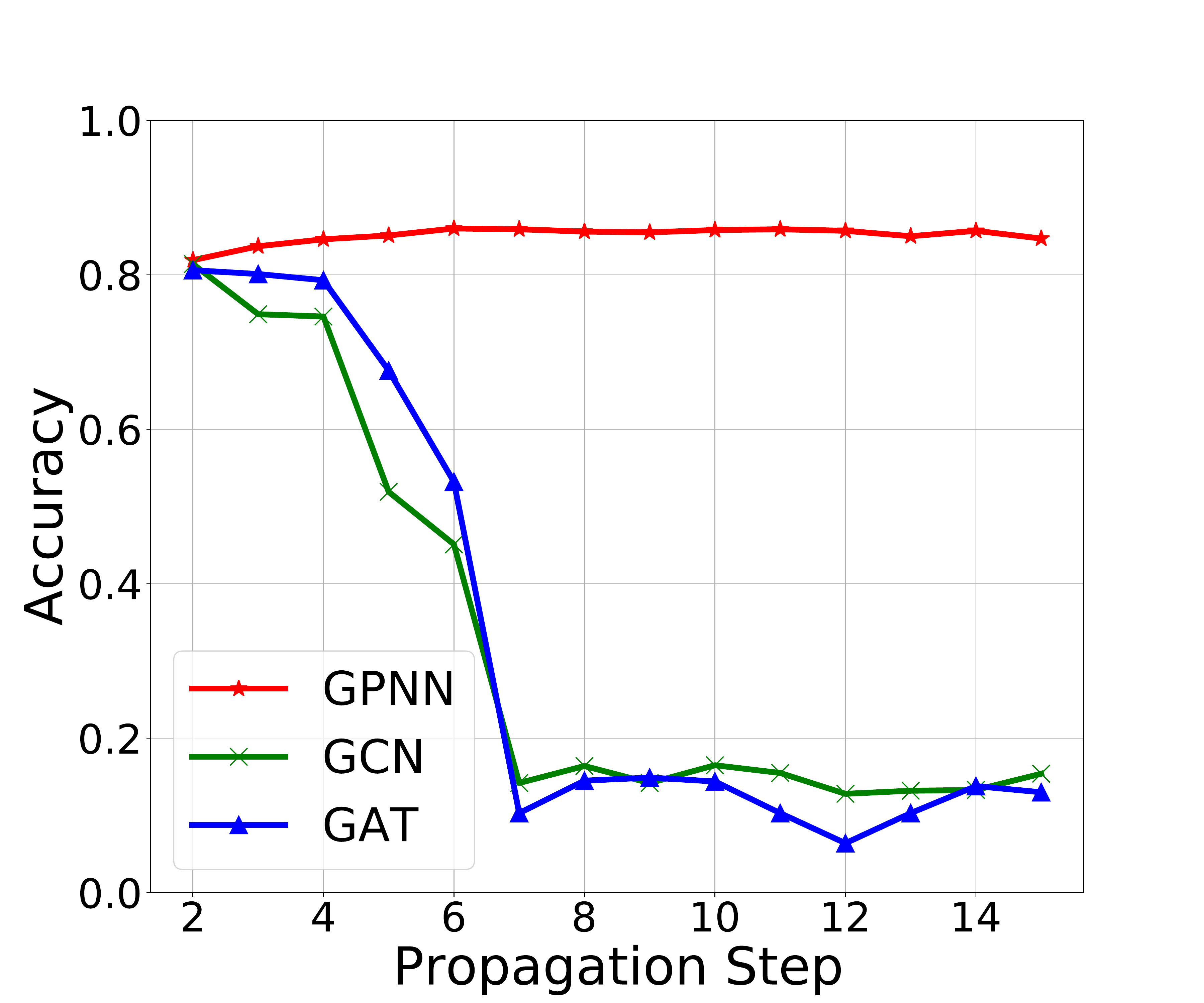}\label{subfig:oversmothing_accuracy}}
    \caption{Over-smoothing analysis on Cora in terms of MADGap and accuracy w.r.t. different propagation steps.}
    \label{fig:experiment_over_smothing_analysis}
\end{figure}
As is shown in Figure~\ref{fig:experiment_over_smothing_analysis}, when models are shallow, although the MADGap value of GPNN is smaller than that of GCN and GAT, our model achieves a higher classification accuracy on Cora. This phenomenon indicates that moderate smoothness of node embeddings can improve performance on classification. Meanwhile, when the propagation step increases to 6, the MADGap and classification accuracy of GCN and GAT drop rapidly while these two metrics of GPNN remain stable and get promoted slightly. We can conclude that GPNN is robust to over-smoothing, which we attribute to the well-designed parameters sharing MLP partner and the consistency contrastive loss. Additionally, by removing transformation in the middle graph convolution layers, GPNN significantly raises the lower bound of the Dirichlet energy which indicates a clear mitigation of the over-smoothing issue. Detailed analysis is provided in the Appendix.

\subsection{Ablation Study}
We investigate the contributions of each component from the perspective of improving model performance and relieving over-smoothing.

\paragraph{Graph Partner and Rectification Loss}
In this part, we take a deeper insight into the contributions of both the proposed graph partner and the rectification loss. 
Since both of these two components cannot directly influence the learning of the embedding matrix $\mathbf{H}^{(l+1)}$,
\begin{figure}
    \centering
    \subfigure[Cora]{\includegraphics[width=0.48\columnwidth,scale=1.2]{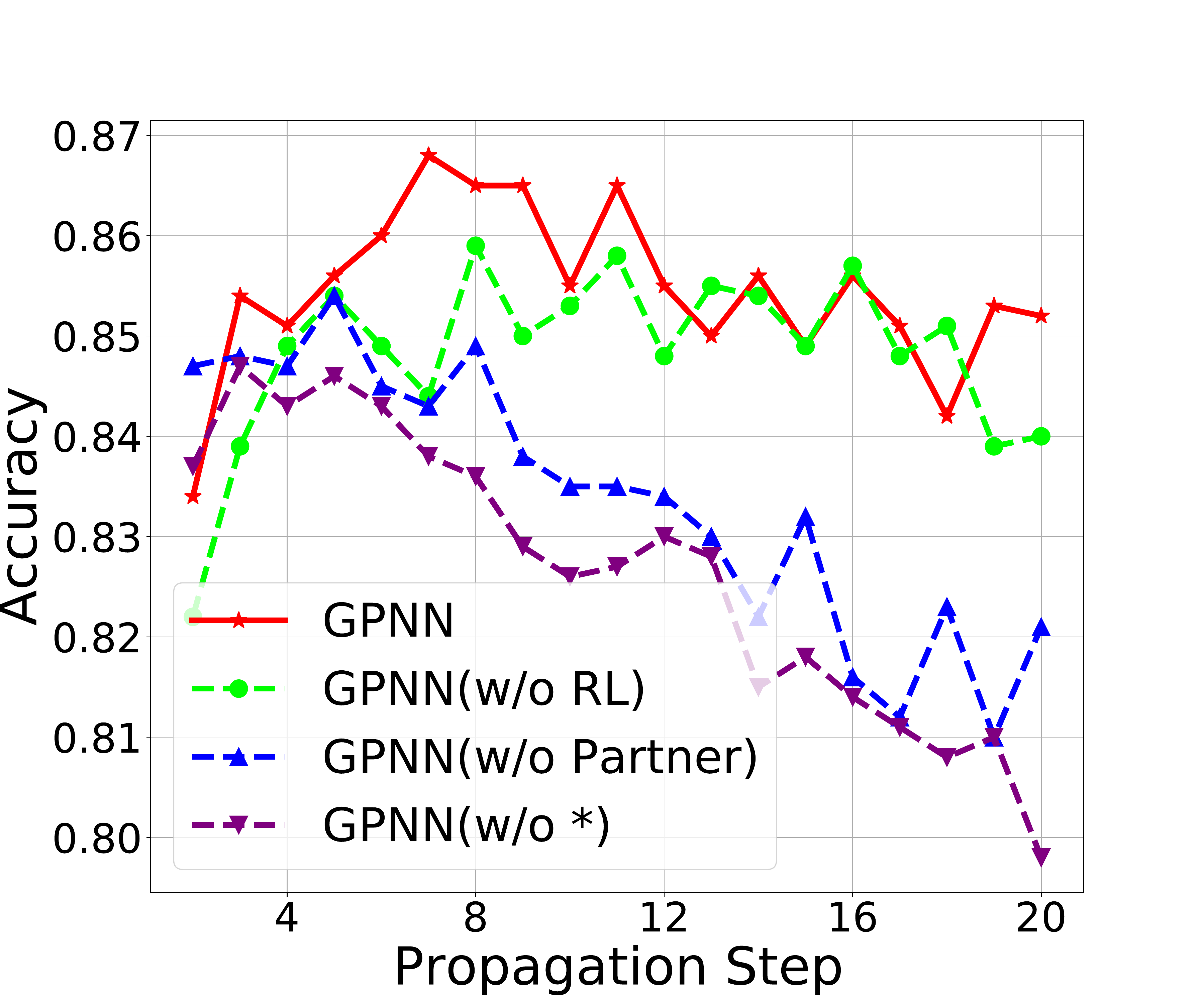}}
    \subfigure[Citeseer]{\includegraphics[width=0.48\columnwidth,scale=1.2]{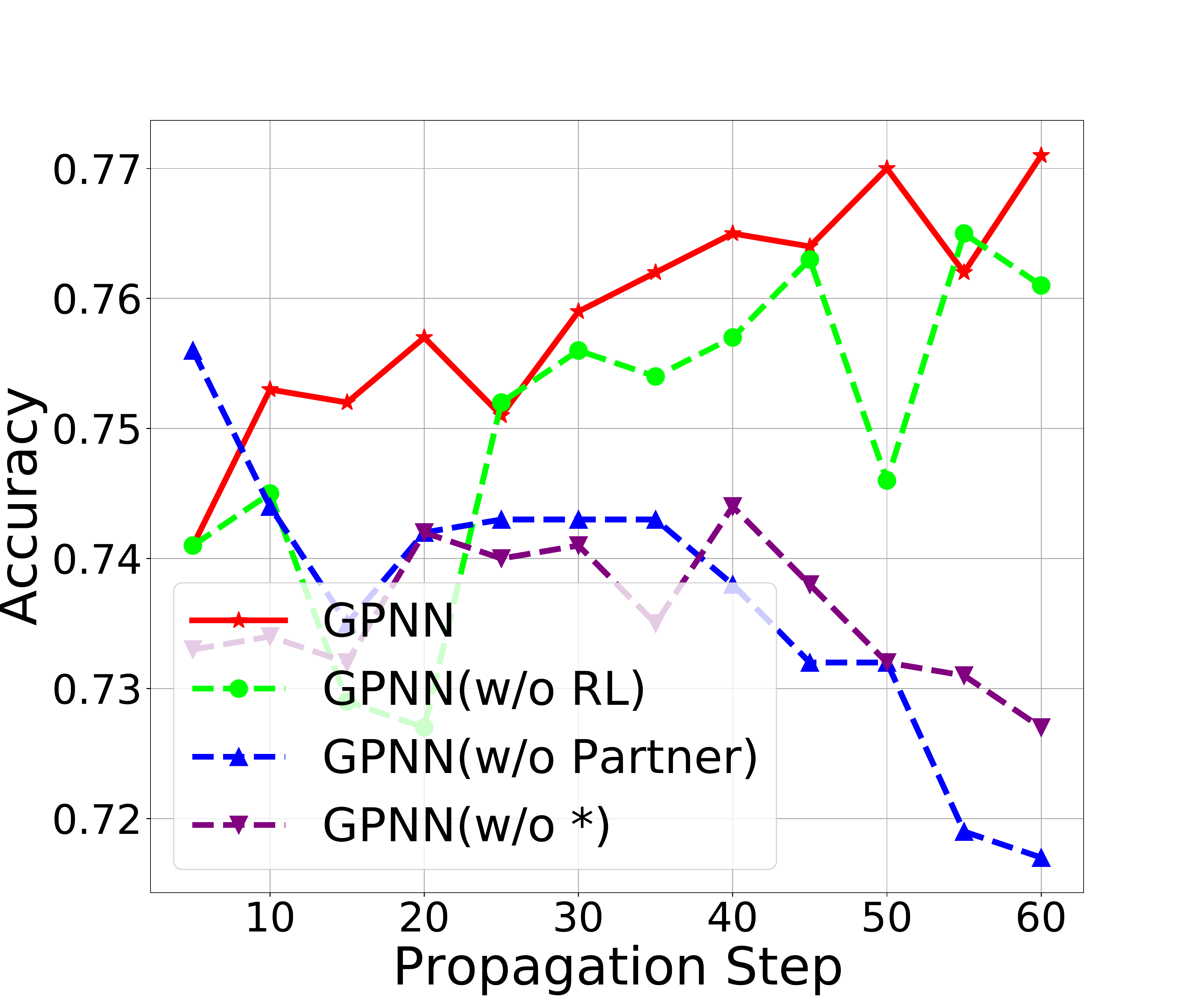}}
    \caption{Accuracy w.r.t. to different propagation steps on Cora and Citeseer. GPNN(w/o) RL corresponds to GPNN without rectification loss. GPNN(w/o Partner) represents that only use GCN for prediction. GPNN(w/o *) corresponds to GPNN without both graph Partner and rectification loss.}
    \label{fig:experiment_propagation_steps}
\end{figure}
we analyze their effects from how the performance changes when the propagation step continuously increases.
From Figure\ref{fig:experiment_propagation_steps}, we observe that the blue line declines faster when the propagation step increases while the red line stays more stable and a huge gap exists between them. This result suggests that graph partner is especially useful for tackling over-smoothing and improving accuracy. We attribute this quality to the essence of MLP for preserving the feature of the node itself. Furthermore, the rectification loss has a slight positive effect on performance but it seems that rectification loss does not help relieve over-smoothing. In practice, when the training process reaches a certain number of epochs, the rectification loss remains almost unchanged, which may mean that the model has reached a balance point for boosting the two branches learning from each other and classifying nodes.

\begin{figure}
    \centering
    \subfigure[Dirichlet Energy] {\includegraphics[width=0.48\columnwidth,scale=1.2]{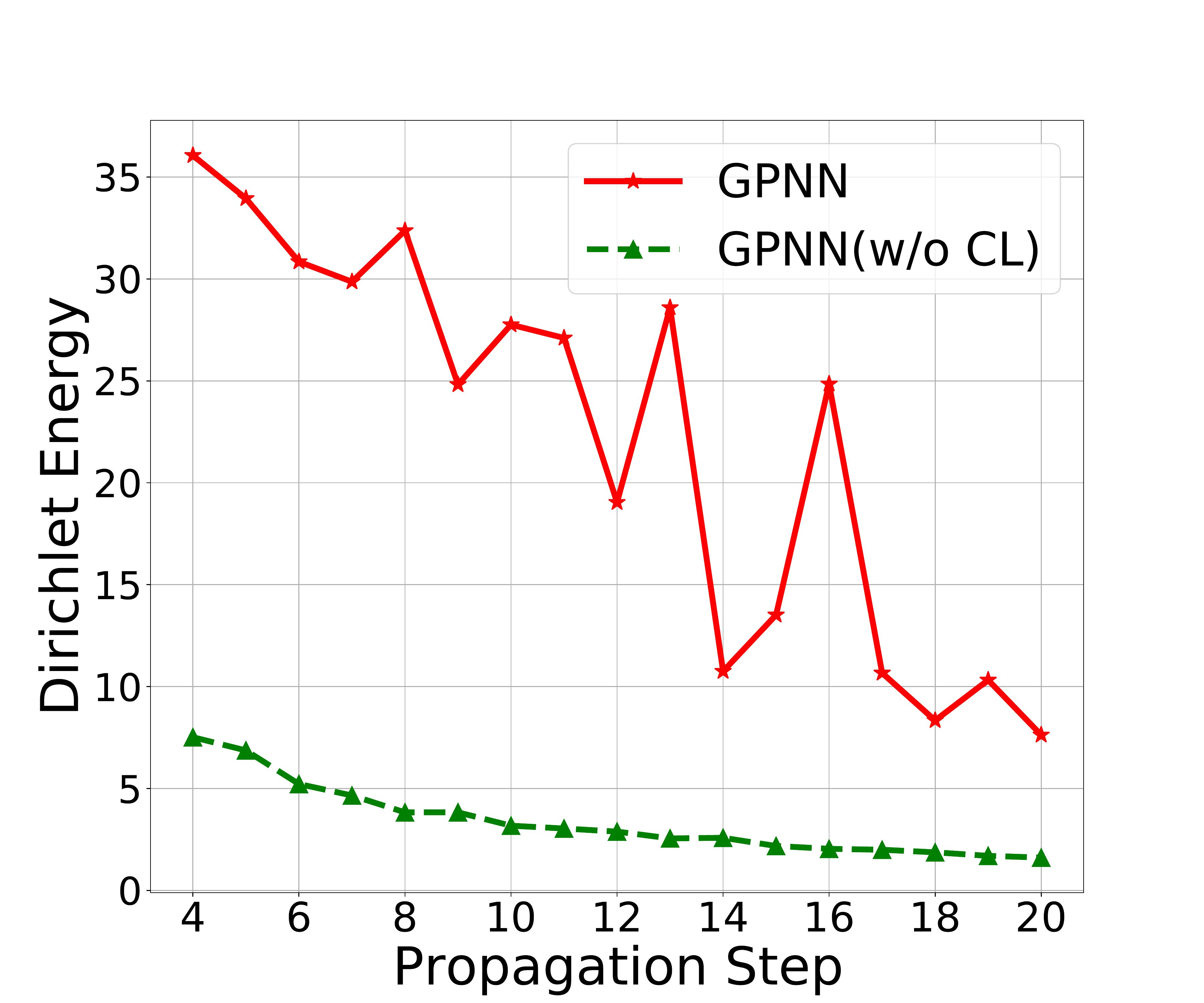}\label{fig:3a}}
    \subfigure[Accuracy]
    {\includegraphics[width=0.48\columnwidth,scale=1.2]{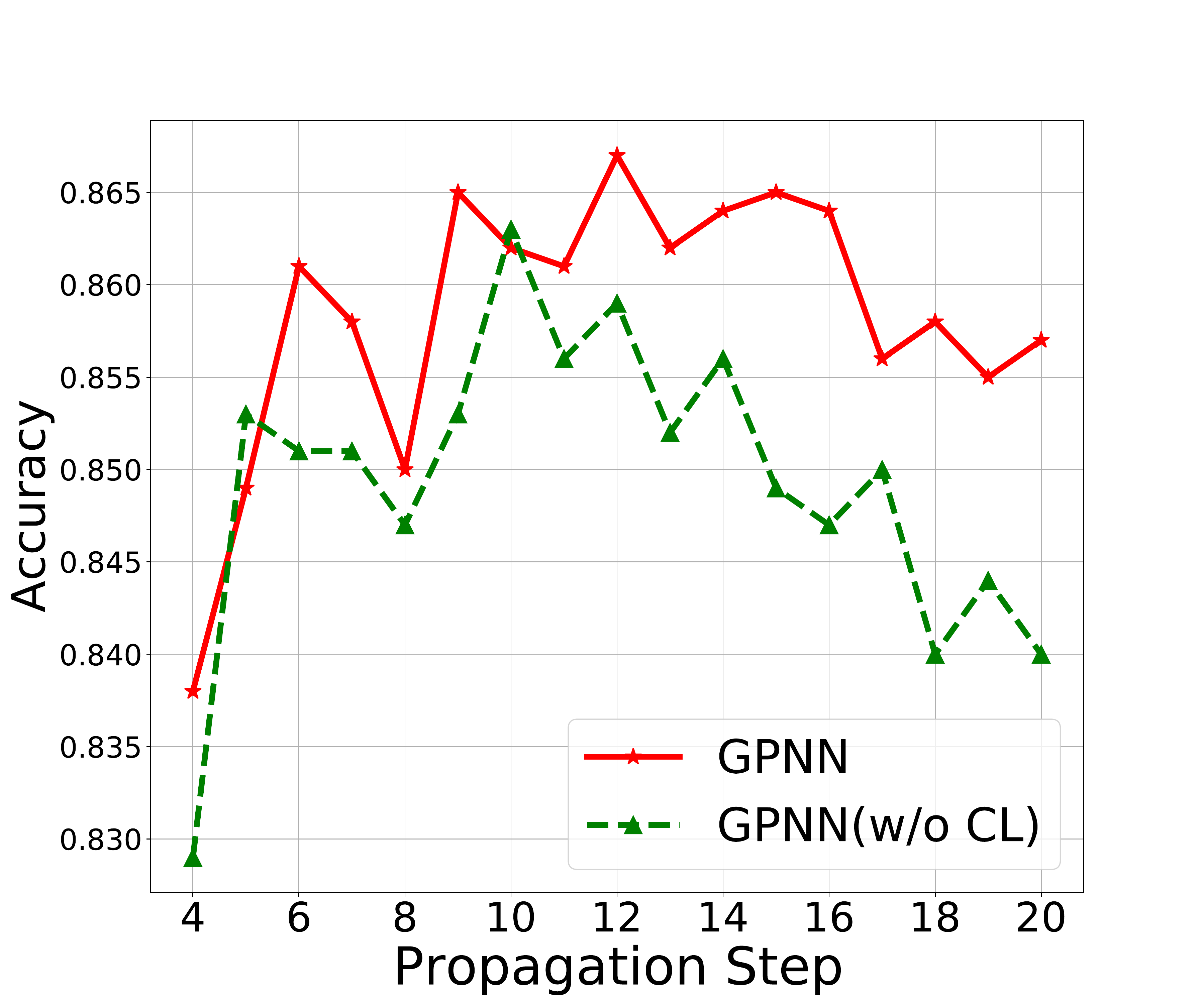}}
    \caption{Dirichlet energy of $\mathbf H^{(l+1)}$ and accuracy on Cora w.r.t. different propagation steps. GPNN(w/o CL) corresponds to GPNN without contrastive loss.}
    \label{fig:experiment_dirichlet}
\end{figure}

\paragraph{Consistency Contrastive Loss}
We investigate the contributions of the consistency contrastive loss by analyzing the effect it puts upon $\mathbf H^{(l+1)}$ since it's directly applied on $\mathbf H^{(l+1)}$. Recall that an extremely small Dirichlet energy reveals that the model is severely suffering from over-smoothing. We make two observations from Figure \ref{fig:experiment_dirichlet}: 1) The red line is consistently higher than the green line in Figure \ref{fig:3a}, which indicates that the proposed consistency contrastive loss significantly raises the Dirichlet energy of $\mathbf{H}^{(l+1)}$, thus making the nodes better separated in the embedding space.
2) As the number of layers increases, the Dirichlet energy gradually drops, but the accuracy first rises and then declines. This further confirms the experimental conclusion we have drawn that proper smoothness of node embeddings can improve performance.

\begin{table}[]
\centering
\caption{Average cosine similarity and average same-label rate statistics for original edges and newly added edges by graph enhancement with $t=0.4$. We only list the statistics on Cora, Pubmed, and Coauthor CS due to space limitation. The rest is provided in the Appendix. Rate corresponds to average same-label rate, similarity corresponds to average cosine similarity, - represents the results of the originally connected node pairs and * represents the results of the newly connected node pairs. Acc. corresponds to accuracy. Acc.(w/o En) corresponds to accuracy of GPNN without graph enhancement.}
\label{tab:experiment_enhenced_graph_result}
\begin{tabular}{c|ccc}
\midrule
\midrule
Metric          & Cora & Pubmed & Coauthor CS \\
          \midrule
Rate-    & 81.0   & \textbf{80.2}     & 80.8   \\
Rate*   & \textbf{90.1} & 78.3     & \textbf{90.2}   \\
\midrule
Similarity-  & 0.17      &0.27          &0.40\\
Similarity* &\textbf{0.53}     &\textbf{0.50}            &\textbf{0.49}\\
\midrule    
Acc.(w/o En) &85.5 ± 0.5 &80.4 ± 0.4    &92.2 ± 0.3\\
Acc.     &\textbf{86.0 ± 0.4}  &\textbf{80.8 ± 0.4}    &\textbf{93.1 ± 0.5}\\
\midrule
\end{tabular}
\end{table}

\paragraph{Impact of Graph Enhancement}
The enhanced graph encodes both structure and attribute information and makes the connections (edges) denser. We empirically analyze its impact on learning representations from the perspective of connection quality and model performance. We adopt the average cosine similarity and average same-label rate over all the connected node pairs to quantitatively measure the quality of connections. The same-label rate is defined as the proportion of nodes of the same class in all the first-order neighbors. \cite{DBLP:conf/aaai/ChenLLLZS20} conducted experiments to prove that even with the same model and propagation step, nodes with higher same-label rate (which they call information-to-noise ratio) generally have higher prediction accuracy. Table \ref{tab:experiment_enhenced_graph_result} lists the average cosine similarity and average same-label rates of the original connections and new connections added by graph enhancement on Cora, Pubmed, and Coauthor CS. As one can see, new connections have high quality in terms of similarity and same-label rate, which indicates that graph enhancement can detect new connections that generally have better quality compared with the original edges. Graph enhancement is especially useful for Coauthor CS and Citeseer, and not very beneficial to other datasets. We speculate that the improvement of performance is highly correlated to the dataset itself. For example, graph enhancement can greatly improve the quality and amount of the connections for Coauthor CS and thus leads to a significant performance improvement.

\begin{table}[]
\centering
\caption{Efficiency comparison of GPNN with GCNII (wall time, in secs., average values of 10 runs). The results include all the time overhead (including the computation time of graph enhancement).}
\label{tab:efficiency}
\begin{tabular}{c|cccc}
\midrule
\midrule
Dataset     & GCNII & GPNN \\
\midrule
Cora        & 94.9  & 10.9($8.71\times$)  \\
CiteSeer    &48.8   & 31.5($1.55\times$)  \\
PubMed      & 20.7 &  37.3($0.55\times$) \\
Coauthor CS &171.6  &  65.4($2.62\times$)\\
A-Computer  &79.8  &  10.1($7.90\times$) \\
A-Photo     & 152.7  &  9.7($15.74\times$) \\
\midrule
\end{tabular}
\end{table}

\subsection{Training Efficiency Comparison}
We compare GPNN with the most competitive deep model GCNII on training efficiency. For a fair comparison, we implement GCNII with the official code (\url{https://github.com/chennnM/GCNII}) and adopt their original hyperparameters for Cora, Citeseer, and Pubmed, while tuning better hyperparameters for the other three datasets since no records are available in the original paper. We conduct experiments with a machine running Ubuntu 20.04LTS with two Intel Xeon Gold 6240L CPUs, one 32GB NVIDIA V100 GPU, and 256GB Memory. From Table \ref{tab:efficiency} we notice that GPNN takes less time to converge than that of GCNII on 5 out of 6 datasets. In particular, GPNN has a large advantage over GCNII with speedup $15.74\times$ on Amazon-Photo. Another worth noting point is that GPNN has much fewer parameters compared with GCNII, which means less memory usage.
\section{Conclusion}
In this paper, we propose GPNN, a simple method for training deep models. GPNN consists of a de-parameterized GCN and a parameters-sharing MLP partner. We further leverage contrastive learning and consistency regularization to ensure the consistency between representations of different times and of different learning branches. we also propose a graph enhancement technique for improving the connection quality. Experimental results show that GPNN outperforms the state-of-the-art baselines on various semi-supervised node classification tasks. Besides, extensive experiments are conducted to investigate the contributions of each component and show efficiency advantage over GCNII, the most powerful deep model in the literature.

\bibliographystyle{named}
\bibliography{ijcai22}

\end{document}